\documentclass[10pt,twocolumn,letterpaper]{article}

\usepackage{cvpr}              

\usepackage{graphicx}
\usepackage{amsmath}
\usepackage{amssymb}
\usepackage{booktabs}
\usepackage{bm}
\usepackage{comment}
\usepackage[accsupp]{axessibility}
\usepackage{xcolor}
\usepackage{xspace}

\newcommand{\approach}{BH-NeRF\xspace}

\newcommand{\sgra}{SgrA$^*$}
\newcommand{\emission}{e}

\newcommand{\xyz}{\left(x,y,z\right)}
\newcommand{\bfx}{{\bf x}}

\newcommand{\rotation}{{\bf R}}
\newcommand{\angularvelocity}{\omega}
\newcommand{\rotaxis}{{\bm \xi}}
\newcommand{\rotangle}{\phi}
\newcommand{\measurements}{{\bf y}}

\newcommand{\fourier}{{\bf F}_t}
\newcommand{\image}{{\bf I}}
\newcommand{\pixel}{p}
\newcommand{\raypath}{\Gamma}
\newcommand{\intweight}{\Delta s}
\newcommand{\measnoise}{{\bm \varepsilon}}
\newcommand{\transpose}{{\operatorname{T}}}
\newcommand{\mlp}{\operatorname{MLP}}
\newcommand{\riaf}{\operatorname{RIAF}}
\newcommand{\netparams}{\bm \theta}
\newcommand{\rayparam}{s}
\newcommand{\todo}[1]{{\color{red} TODO: #1}}

\usepackage[pagebackref,breaklinks,colorlinks]{hyperref}

\usepackage[capitalize]{cleveref}
\crefname{section}{Sec.}{Secs.}
\Crefname{section}{Section}{Sections}
\Crefname{table}{Table}{Tables}
\crefname{table}{Tab.}{Tabs.}

\def\cvprPaperID{4415} 
\def\confName{CVPR}
\def\confYear{2022}

\newcommand\blfootnote[1]{%
  \begingroup
  \renewcommand\thefootnote{}\footnote{#1}%
  \addtocounter{footnote}{-1}%
  \endgroup
}

\begin{document}

\title{Gravitationally Lensed Black Hole Emission Tomography}




\author{Aviad Levis$^{1*}$,\quad Pratul P.~Srinivasan$^{2*}$,\quad Andrew A.~Chael$^{3 \dagger}$,\quad Ren Ng$^{4}$,\quad Katherine L.~Bouman$^{1}$ \vspace{0.2cm} \\ 
$^1$California Institute of Technology \quad $^2$Google Research
\quad $^3$Princeton \quad $^4$UC Berkeley \vspace{0.2cm} \\
}

\maketitle

\blfootnote{$^*$ Authors contributed equally to this work.}
\blfootnote{$^\dagger$ NASA Hubble Fellowship Program, Einstein Fellow.}

\vspace{-0.7cm}
\begin{abstract}
\vspace{-0.2cm}
Measurements from the Event Horizon Telescope enabled the visualization of light emission around a black hole for the first time. So far, these measurements have been used to recover a 2D image under the assumption that the emission field is static over the period of acquisition. In this work, we propose \approach, a novel tomography approach that leverages gravitational lensing to recover the continuous 3D emission field near a black hole. Compared to other 3D reconstruction or tomography settings, this task poses two significant challenges: first, rays near black holes follow curved paths dictated by general relativity, and second, we only observe measurements from a single viewpoint. Our method captures the unknown emission field using a continuous volumetric function parameterized by a coordinate-based neural network, and uses knowledge of Keplerian orbital dynamics to establish correspondence between 3D points over time. Together, these enable \approach to recover accurate 3D emission fields, even in challenging situations with sparse measurements and uncertain orbital dynamics. 
This work takes the first steps in showing how future measurements from the Event Horizon Telescope could be used to recover evolving 3D emission around the supermassive black hole in our Galactic center. 
\end{abstract}

\vspace{-0.25cm}
\section{Introduction}
\label{sec:intro}

\begin{figure}
	\centering \includegraphics[width=\linewidth]{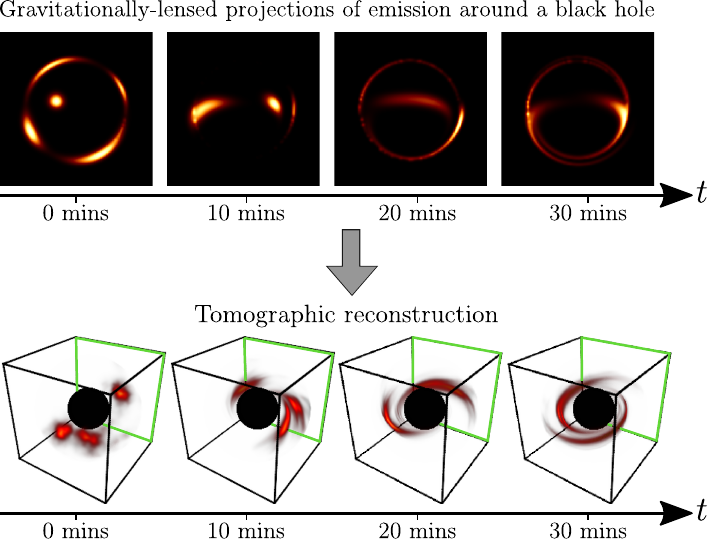}
	\caption{We formulate and present an approach for a novel tomography: recovering the emission distribution of flares in orbit around a black hole using observations captured from a single viewpoint over time.
	This is achieved by leveraging expected physics around a black hole: gravitational lensing and emission flow modeled by Keplerian orbital dynamics. 
	In this example, observations are simulated telescope measurements that place a sparse set of constraints on the Fourier components of the 2D images images over time (top row); The image formation model accounts for ray trajectories that are curved due to gravitational lensing. The orientation of the observed projection plane is indicated in green. Our method, \approach, recovers a representation of the underlying 3D emission from the sparse telescope measurements.}
	\label{fig:teaser}
\end{figure}

The Event Horizon Telescope (EHT), which recently captured the first image of a black hole, opens the door to a new era in black hole research~\cite{akiyama2019firstIV}. We are now able to image the direct environment around a black hole and resolve image features on the scale of an event horizon. This unprecedented resolution is achieved by linking radio telescopes across the globe to synchronously observe an astronomical source~\cite{akiyama2019firstII, bouman2016computational,chael2018interferometric}, a technique known as Very Large Baseline Interferometry~\cite{thompson1999fundamentals}.
Although a black hole itself is not visible, its imprint or ``shadow" on the emission from the surrounding hot gas is what the EHT aims to image. 

The two candidate black holes within reach of the EHT are M87$^*$ and Sagittarius A$^*$ (\sgra). The first black hole image of M87$^*$~\cite{akiyama2019firstIV} was constructed from measurements collected over an entire night, with the underlying assumption that its emission is static throughout acquisition. In contrast, \sgra, the black hole in our Galactic center, is far more dynamic and displays strong flares of emission up to several times a day~\cite{genzel2003near}. A potential explanation for these flares are compact bright emission spots~\cite{broderick2005imaging} in orbit around the black hole, referred to as {\it hot-spots}.

In this work, we formulate and propose an approach for a novel tomography problem to reconstruct these dynamic events (Fig.~\ref{fig:teaser}). In contrast to previous works~\cite{bouman2018reconstructing,levis2021inference, johnson2017dynamical}, which focus on recovering 2D image plane dynamics, our work recovers the underlying dynamic 3D emission distribution (i.e., 4D spatiotemporal volume). However, reconstructing volumetric emission in this setting faces two salient challenges. The first challenge is that light rays which reach the EHT telescopes do not follow straight lines that regularly sample space. Instead, the rays follow curved paths due to the strong gravitational field around black holes. This effect, predicted by general relativity (GR), is known as {\it gravitational lensing} since massive objects, such as black holes, act as lenses and bend light. The second challenge is the EHT only observes from a single fixed viewpoint, as opposed to the multi-view observations acquired in other tomography settings. 

In this paper, we introduce a novel tomography that leverages expected physics around a black hole to recover the time-varying volumetric emission from a single observational direction. In particular, we present a promising solution, \approach, that represents the 3D emission as a continuous function using a coordinate-based neural network and makes use of our knowledge of Keplerian orbit dynamics to relate 3D emission points across time. 

Our approach builds upon recent advances in neural representations of radiance fields~\cite{mildenhall2020nerf} and generalizes them to account for curved ray trajectories. 
Our experiments demonstrate promising first steps towards recovering the 3D dynamics around a black hole, a high-priority task for astronomy research over the upcoming decade~\cite{astro2020}.

\section{Related Work}
\label{sec:prior}

\paragraph{Tomographic 3D Reconstruction}
Tomography aims to recover densities of a medium from lower dimensional projection measurements. In medical imaging, where X-ray computed tomography (CT)~\cite{KakSlaney88} is extensively used for diagnostics, 2D projections from different orientations are used to recover a 3D volume. These CT approaches are a well established technology, partly due to the fact that simple photon-medium interactions result in straight-rays through the medium~\cite{gordon1970algebraic}. 

In many scientific fields, complex interactions result in non-straight light paths (e.g. scattering, refraction, lensing) and non-linear measurement models, rendering tomography a largely open problem for research.
In Earth Science, atmospheric tomography using scattered light gives rise to broken-ray paths and non-linear measurements~\cite{levis2015airborne,  levis2017multiple, levis2020multi}. In Seismology, imaging the interior of the Earth involves resolving refraction and propagation along curved trajectories~\cite{gao2021deepgem}. For optical imaging, a similar effect is caused by a medium with variable refractive index~\cite{atcheson2008time, xiong2021wild}. In Astrophysics, curved light paths caused by gravitational lensing around high density regions in space has been used to constrain the 3D structure of dark matter~\cite{cosmotomography}. Analogously, the strong gravitational field near a black hole causes light paths to curve (refer to Fig.~\ref{fig:ray_illustration}). 

In the context of black hole imaging, the work of Tiede \etal~\cite{tiede2020spacetime} showed recovery of a handful of parameters related to orbiting hot-spots using Monte-Carlo sampling. While limited to a parameterized model, their work showed promise in recovering 3D information from 2D projections.
Our work is inherently different from~\cite{tiede2020spacetime} as it is not limited to a simple parameteric form. Rather, we aim to recover a {\em continuous} unknown 3D volume. 

\vspace{-0.3cm}
\paragraph{Coordinate-based Neural Representations}
A recent trend in computer vision and graphics is to replace traditional discrete representations of geometry, such as triangle meshes and voxel grids, with continuous ``coordinate-based'' neural representations~\cite{chen2019imnet,mescheder2019occupancy,park2019deepsdf}. This approach parameterizes continuous 3D fields with a multilayer perceptron (MLP) that maps from an input 3D coordinate to a representation of scene content at that location~\cite{mildenhall2020nerf}. 

The use of neural representations for rendering unobserved views of scenes was popularized by Neural Radiance Fields (NeRF)~\cite{mildenhall2020nerf}, which outputs the volume density and view-dependent radiance at any continuous 3D point in the scene. Our work is related to extensions of NeRF to dynamic scenes~\cite{gafni2021dynamic,li2021neural,park2021nerfies,peng2021animate,tretschk2021nrnerf}. In dynamic settings, a key challenge is relating 3D points across time so that the scene can be represented by deformations of a canonical 3D representation. Previous methods use priors on deformation sparsity and rigidity~\cite{park2021nerfies,tretschk2021nrnerf}, pre-trained monocular depth estimation~\cite{li2021neural}, or explicit models of human faces and bodies~\cite{gafni2021dynamic,peng2021animate} to relate 3D points across time. In our work, we leverage knowledge of Keplerian orbit dynamics to model the dynamic evolution around a black hole.

We take inspiration from the success of coordinate-based neural representations in other scientific imaging domains, such as cryo-electron microscopy~\cite{zhong2021cryodrgn} and X-ray tomography~\cite{tancik2020fourfeat}, and use it to represent the evolving 3D emission around a black hole as described in the following sections.


\vspace{-0.1cm}
\section{Image \& Measurement Formation Model}
\vspace{-0.1cm}
\label{sec:formation}
In this section, we discuss the three components of our ``forward model", which models measurements observed by the EHT as a function of the emission surrounding a black hole: 1) the emission dynamics that describe the evolution over time, 2) the gravitationally lensed ray tracing that describes how 3D emission is integrated along curved rays to form a 2D image, and 3) the measurement model that describes how EHT measurements relate to these images.

\subsection{Emission Dynamics}
\label{subsec:emission_dynamics}
We represent emission, $\emission \left(t, \bfx \right)$, as a continuous function of time $t$ and 3D coordinate $\bfx = \xyz$. Emission around a black hole can be the result of either inflowing plasma from an accretion disk or outflowing plasma from a jet. Here, we assume the emission originates in the equatorial plane, where the temporal dynamics are modeled by a Keplerian angular velocity $\angularvelocity(r)$ that decreases with distance $r$ from the black hole center~\cite{broderick2006imaging}. In particular, the angular frequency of an orbit at radius $r$ around a black hole with mass $M$ and zero angular momentum is given by:
\begin{equation}
    \angularvelocity(r) = \frac{1}{2 \pi \sqrt{r^3/GM}} \propto r^{-3/2}.
    \label{eq:angmom}
\end{equation}
where $G$ is the gravitational constant. 
The extension of \eqref{eq:angmom} for cases where the black hole angular momentum (i.e., spin) is nonzero is given in the supplemental material~\cite{projectpage}. 

\begin{figure}
	\centering \includegraphics[width=0.9\linewidth]{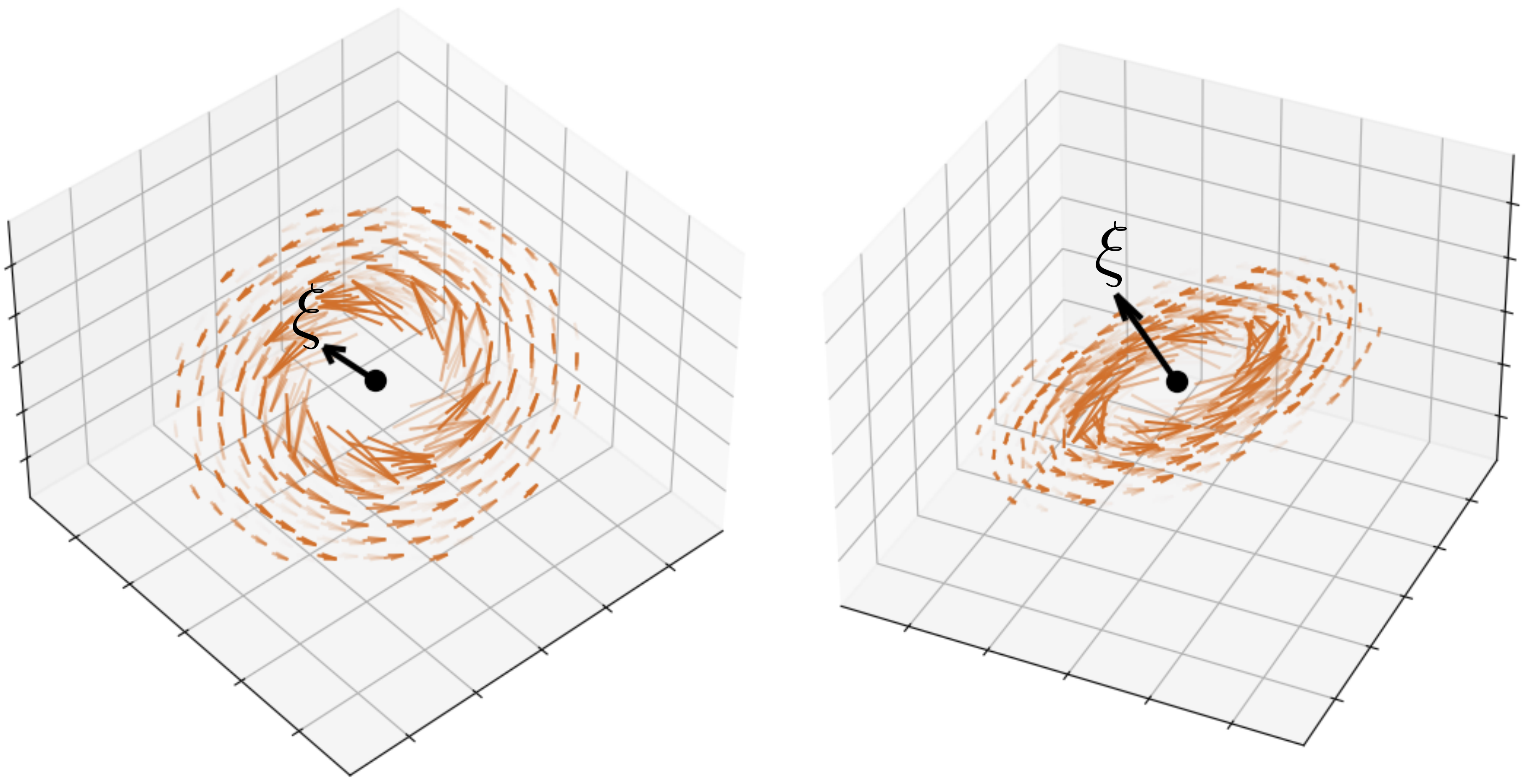}
	\caption{An illustration of a Keplerian velocity field about a rotation axis $\rotaxis$, where the velocity at 3D locations is depicted by arrows whose length signifies the velocity's magnitude. The velocity magnitude decreases with distance, as described by Eq~\eqref{eq:angmom}. Our approach does not assume knowledge of the axis  $\rotaxis$ which is jointly estimated with the underlying 3D emission (Sec.~\ref{subsec:rot_axis}). }
	\label{fig:velocity_field}
\end{figure}

The continuous change in angular velocity with radial distance causes a {\em shearing} of structures  (Fig.~\ref{fig:velocity_field}).
The shearing operation can be represented by a coordinate transformation of a canonical emission at time $t=0$:
\begin{equation}
    \emission \left(t, \bfx \right) = \emission_0 \left( \rotation_{\rotaxis, \rotangle} \bfx \right),
    \label{eq:dynamics}
\end{equation}
where $\rotation_{\rotaxis, \rotangle}$ is a rotation matrix of angle $\rotangle$ about the axis $\rotaxis$\footnote{e.g. as given by the {\em Rodrigues' Rotation Formula}}. The position and time dependent angle $\rotangle$ is given by
\begin{equation}
\rotangle \left(t, r\right) = t \angularvelocity(r) \propto t r^{-3/2}.
\label{eq:angular_velocity}
\end{equation}

\begin{figure}[t]
	\centering \includegraphics[width=0.85\linewidth]{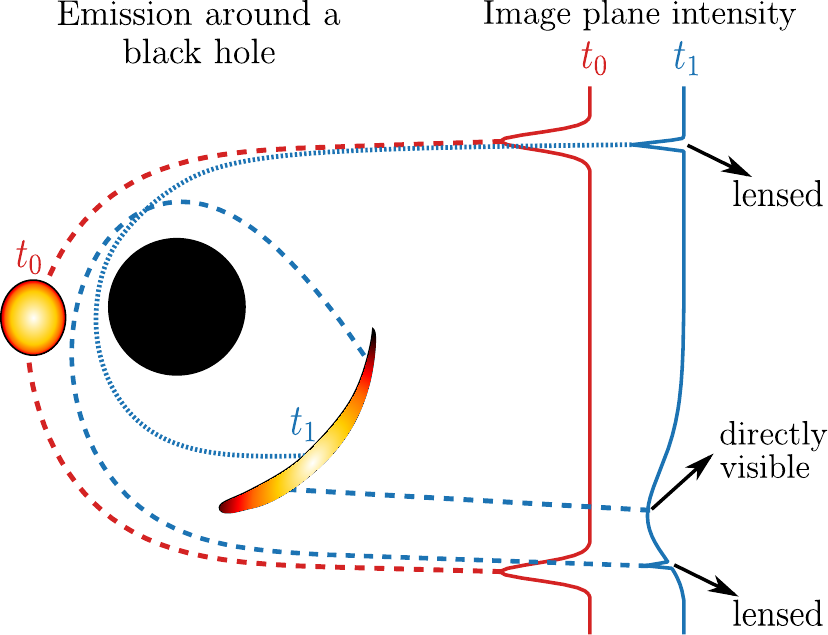}
	\caption{2D Illustration of an emission hot-spot orbiting around a black hole and the resulting 1D image plane projection. As time progresses, $t_0 \rightarrow t_1$, the spot is sheared due to the inner radii moving faster than the outer radii (Sec.~\ref{subsec:emission_dynamics}). The color-coded measurements illustrate the projected image at two different times. Selected rays are traced to highlight interesting image-features caused by the gravitationally curved trajectories.}
	\label{fig:ray_illustration}
\end{figure}

\subsection{Ray Tracing with Gravitational Lensing}

The radiance incident any pixel $\pixel_n(t)$ at time $t$ is given by ``backward'' integration of 3D emission over a ray path, $\raypath_n$, originating at the image plane. For $N {\times} N$ pixels the discretized image plane is defined by the vector
\begin{equation}
\image(t) = \left[\pixel_1(t), ..., \pixel_{N^2}(t)\right].
\end{equation}
The curved ray path in General Relativity (GR) can be described as a 4D (one time and three spatial dimensions) parametric function of the distance along the ray, $\rayparam$. If we know the path $\raypath_n = \left(t(\rayparam), \bfx(\rayparam)\right)$ that ends at the $n$'th pixel, we can compute its intensity by integrating over $\rayparam$:
 \begin{equation}
\pixel_n(t) = \int \limits_{\raypath_n} \! \! \emission \left(t, \bfx \right) {\rm d} \rayparam \approx \sum_{\bfx_i \in \raypath_n}  \! \! \emission \left(t, \bfx_i \right) \intweight_i. 
\label{eq:pixel_values}
\end{equation}
Here $\intweight_i$ is the integration weight associated with the $i$-th sample along the path $\raypath_n$ and can be thought of as the distance along the ray's path taken in step $i$. In contrast to standard ray tracing in Euclidean space, the black hole's gravitational field causes light ray trajectories to curve, and these curved ray paths are computed by integrating a differential equation~\cite{geokerr,grallalupsasca}; we provide explicit equations in the supplementary material~\cite{projectpage}. Note that since in this work we assume ray paths are fixed according to the mass and spin of the black hole, pre-computed trajectories are used throughout the tomographic optimization.


In Eq.~\eqref{eq:pixel_values}, we make three simplifying assumptions. First, we assume that the attenuation (due to absorption and scattering) of light by the emitting gas is negligible. Attenuation is important for general astrophysical images but can be neglected for EHT images of black holes Sgr A* and M87~\cite{paperV}.
Second, although gas is moving at relativistic velocities, causing emission to change as the light ray propagates at a finite speed, we do not account for this~\cite{ipole,grtrans,GRay}.
Finally, we do not account for relativistic redshift, which decreases the emission when material is deep in the gravitational field or moving away from the observer.

Figure \ref{fig:ray_illustration} illustrates how an emission hot-spot behind the black hole is ``lensed'' onto the image plane as a consequence of the bending of light rays. In the figure, selected rays are illustrated to showcase interesting image features. A key image feature predicted by GR is bright ``photon rings" produced by light rays that traverse full orbits around the black hole before reaching the sensor~\cite{johnson2020universal}. 

\begin{figure}[t]
	\centering \includegraphics[width=\linewidth]{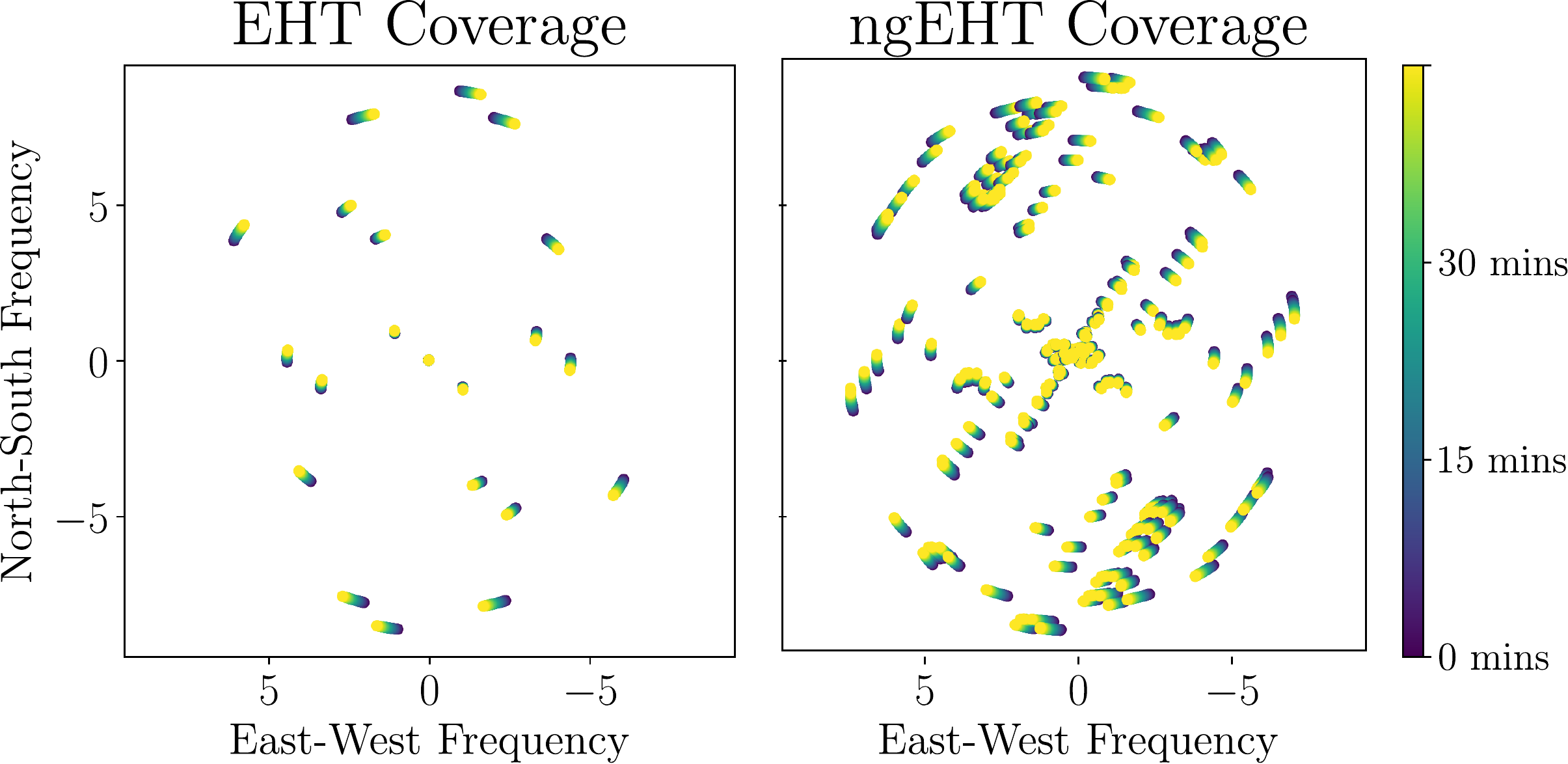}
	\caption{EHT spatial frequency sampling pattern as a function of time. Due to the Earth rotation, EHT telescopes probe different frequencies at different times. The EHT coverage (left) includes existing telescopes whereas the ngEHT~\cite{raymond2021evaluation} also includes candidate telescope sites, resulting in a denser coverage (right). 
	The observation time was chosen to capture $\sim$1 full orbit around \sgra for emission at distances used in our simulations (Sec.~\ref{sec:results}).}
	\label{fig:eht_sampling}
	\vspace{-0.15in}
\end{figure}

\subsection{Event Horizon Telescope Measurements}
\label{subsec:eht}
The Event Horizon Telescope (EHT) records simultaneous radio signals from the black hole source at telescopes located around the globe. The correlation of the recorded radio signals from telescope pairs results in a measurement that probes a single spatial frequency of the underlying 2D image~\cite{thompson1999fundamentals}. The spatial frequency measured is proportional to the projected baselines between the telescopes: short baselines probe low spatial frequencies and long baselines probe high spatial frequencies. 
These measurements, known as {\em complex visibilities}, can be represented as:
\begin{equation}
    \measurements(t) = \fourier \image(t)  +  \measnoise.
    \label{eq:measurements}
\end{equation}
Here, $\fourier$ is the 2D discrete-time Fourier transform (DTFT) matrix containing frequency components sampled by the EHT. The measurement {\it thermal noise}, $\measnoise {=} (\epsilon_1,...,\epsilon_K)^\top$, is Gaussian distributed according to $\epsilon_k \sim \mathcal{N}\left( 0, \sigma^2_k\right)$, where $\sigma_k$ is related to the sensitivity of the telescope pair~\cite{thompson1999fundamentals}. In reality, additional noise sources exist due to the atmosphere and instrumental error~\cite{akiyama2019firstIII}, but in this work we limit ourselves to the fundamental telescope sensitivity.

Figure \ref{fig:eht_sampling} shows the EHT frequency sampling coverage as a function of time during the night of April 7, 2017. The EHT array includes the existing telescopes which were used during the 2017 campaign. The next-generation EHT (ngEHT)~\cite{raymond2021evaluation} array includes additional candidate telescope sites that have been proposed for future EHT observations. These telescope sites, chosen according to favorable atmospheric transmittance and geographic location~\cite{raymond2021evaluation}, yield denser frequency coverage.


\begin{figure*}[t]
	\centering \includegraphics[width=\linewidth]{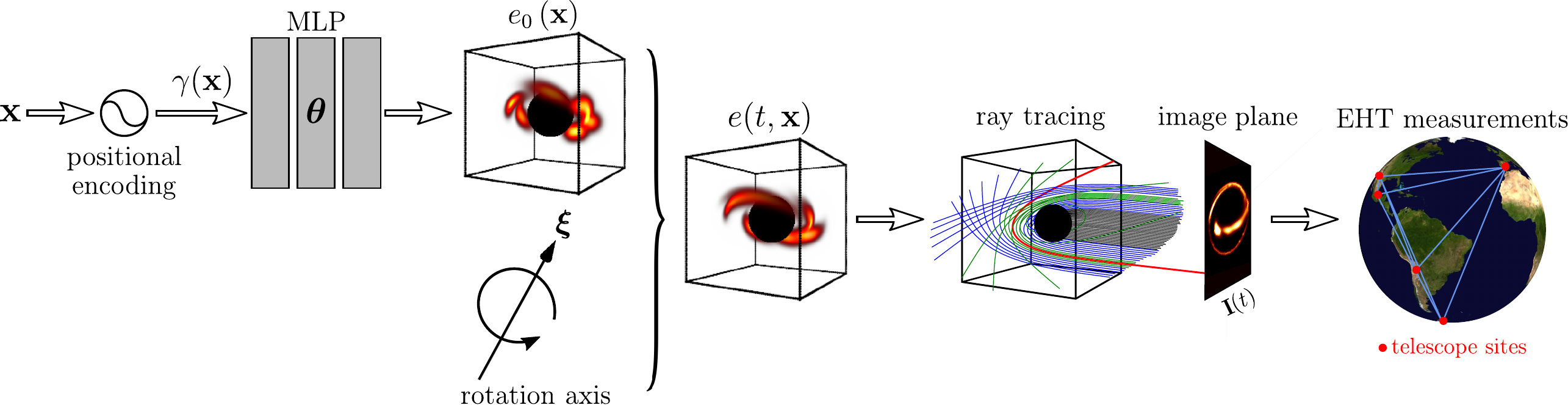}
	\caption{Block diagram of our proposed tomography approach. We model an initial 3D emission $\emission_0$ as a continuous function using an MLP (parameterized by $\netparams$). The input to the network are 3D coordinates $\bfx$ (transformed by a positional encoding--see Sec.~\ref{subsec:emission_MLP}). The orbit dynamics, parameterized by a rotation axis $\rotaxis$, dictate how the initial emission evolves over time $\emission_0 (\bfx) \rightarrow e(t, \bfx)$ (Sec.~\ref{subsec:emission_dynamics}).
	\approach solves for $\netparams$ and $\rotaxis$ jointly using a physically motivated loss that accounts for material orbit and gravitational lensing effects.
	Telescope measurements are used to constrain the optimization over $\netparams$ and $\rotaxis$.
	}
	\vspace{-0.1in}
	\label{fig:approach}
\end{figure*}
 
\section{\approach: {\normalsize Black Hole Neural Radiance Fields}}
\label{sec:tomography}
In this section we discuss how the measurement model (detailed in Sec.~\ref{sec:formation}) is used to supervise a coordinate-based neural network reconstruction of the 3D emission around a black-hole. Figure \ref{fig:approach} gives a block diagram overview. 

\subsection{Representing Emission as an Neural Network}
\label{subsec:emission_MLP}
We take inspiration from NeRF~\cite{mildenhall2020nerf}, parameterizing the 3D volume at time $t=0$ with the weights, $\netparams$, of an MLP neural network. This MLP takes continuously-valued coordinates $\bfx$ as input, and outputs the corresponding emission:
\begin{equation}
    \emission_0 \left(\bfx \right) = \mlp_{\netparams}(\gamma(\bfx))
    \label{eq:MLP}
\end{equation}
where $\gamma(\bfx)$ is a positional encoding of the input coordinates that projects each input coordinate onto a set of sinusoids with exponentially-increasing frequencies:
\begin{equation}
    \gamma(\bfx) = \Big[ \sin(\bfx), \cos(\bfx), \ldots, \sin\!\big(2^{L-1} \bfx\big), \cos\!\big(2^{L-1} \bfx\big) \Big]^\transpose \,.
\end{equation}
This positional encoding controls the underlying interpolation kernel used by the MLP, where the parameter $L$ determines the bandwidth of the interpolation kernel~\cite{tancik2020fourfeat}.

In our experiments, we use an MLP with 4 layers, where each layer is 128 units wide and uses ReLU activations. We use a maximum positional encoding degree of $L=3$. The low degree of $L$ is suitable for volumetric emission fields which are naturally smooth (see Suppl.~\cite{projectpage}). Moreover, EHT measurements observe a limited range of frequencies, thus, it is important to use a positional encoding degree that does not introduce spurious high frequencies that cannot be supervised by the measurements.

\subsection{Rotation Axis Estimation}
\label{subsec:rot_axis}
 \begin{figure}[t]
 	\centering \includegraphics[width=\linewidth]{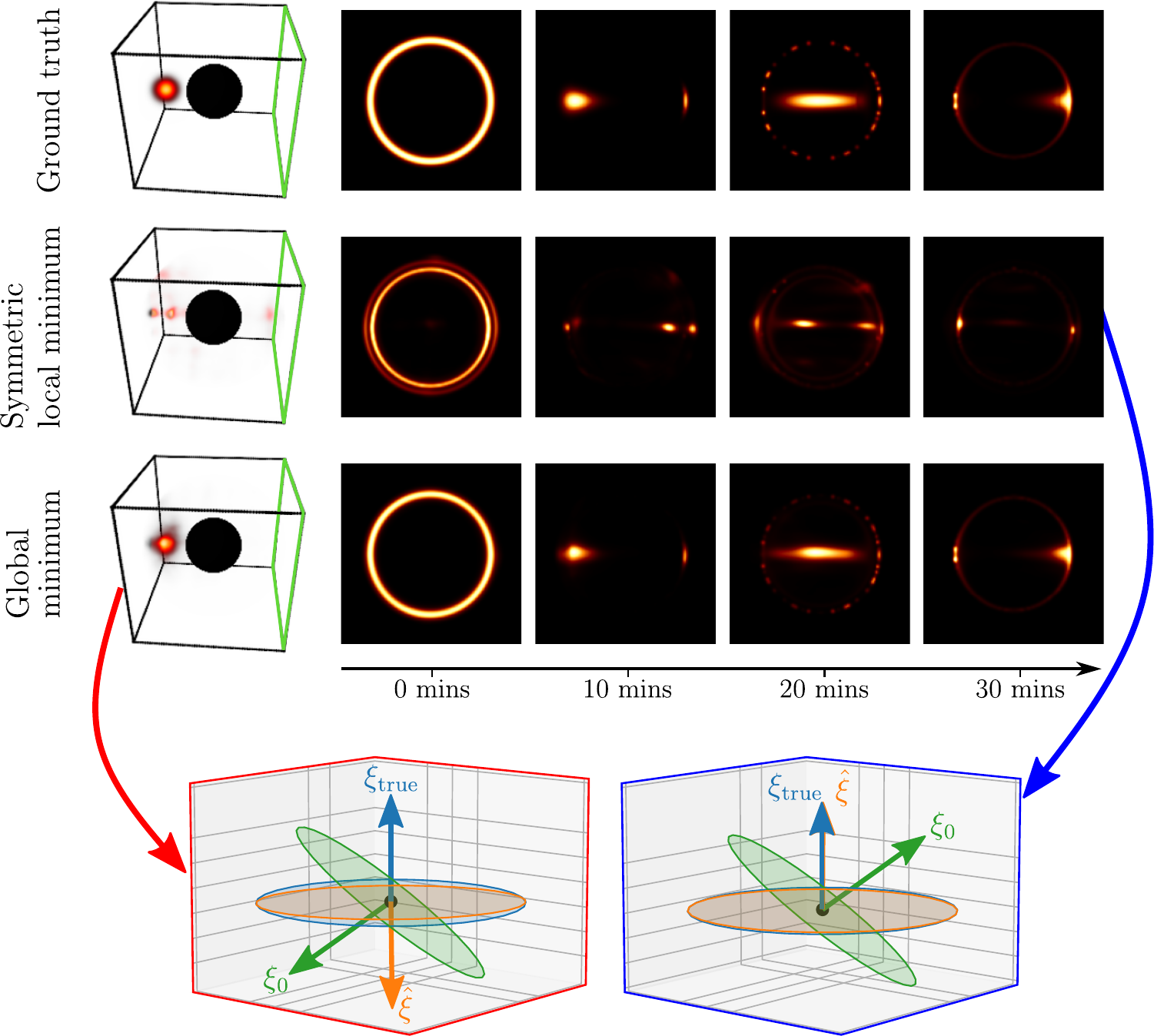}
 	\caption{Symmetry induces a local minimum at the opposite rotation direction. The top row shows the ground truth emission of a single hot-spot on the equatorial plane directly behind the black hole. Top row images are rendered with a rotation axis $\rotaxis_{\rm true}=(0,0,1)$. When initializing $\rotaxis_0$ with a random vector in the southern hemisphere, the converged solution finds a local minimum. Simply initializing the rotation axis to the opposite sign vector (in the northern hemisphere) identifies the global minimum and the hot-spot emission is recovered accurately.}
 	\vspace{-0.2in}
	\label{fig:local_minimum}
\end{figure}
In the previous section we described how we model the 3D emission as a continuous function using a neural network. To recover the unknown emission from measurements, our approach relies on Keplerian motion to model the orbit and evolution over time (Sec.~\ref{subsec:emission_dynamics}). We consider a case where the angular velocity is well modeled by Eq.~\eqref{eq:angular_velocity}, however, the rotation axis $\rotaxis$ (Eq.~\ref{eq:dynamics}) is unknown. 

For \sgra, past studies have attempted to constrain the rotation axis~\cite{broderick2016modeling}. However, these constraints have large uncertainty due to measurement and modeling error. Therefore, our approach estimates an unknown axis jointly with the 3D emission (Fig.~\ref{fig:approach}). The estimated values from other observations may be able to serve as a good prior or initialization for the optimization problem.

Since the forward model is fully differentiable, gradients can be back-propagated to directly optimize the unknown rotation axis, represented as a normalized 3-vector, $\rotaxis$. As highlighted in Fig.~\ref{fig:local_minimum}, we identify a persistent local minimum that occurs at the symmetric axis, equal to the negative of the true rotation axis. This can be intuitively explained: even without temporal aliasing, two opposing orbits share a common rotation plane. Thus, the projected emission results in a common ``line" formed on the image plane, that results in a lower loss than a perturbed rotation axes. This local minimum exists in all presented experiments. 
To avoid this local minimum, we initialize the rotation axis with a random unit vector drawn from the unit sphere and its symmetric counterpart from the opposing hemisphere. We present the resulting emission with the lowest converged loss. 




\subsection{Optimization}
\label{subsec:optimization}
We now detail how all the the components described in Secs.~\ref{sec:formation}, \ref{sec:tomography} are integrated in the full optimization problem. 
The optimization is solved by minimizing a loss function over network parameters $\netparams$ and rotation axis $\rotaxis$: 
\begin{equation}
    \mathcal{L}(\netparams, \rotaxis) = \sum \limits_t \left\|\measurements(t) - \fourier \image_{\netparams, \rotaxis}(t) \right\|^2_{{\bf \Sigma}},
    \label{eq:loss}
\end{equation}
Here, $t$ denotes discrete observational time frames and $\|{\bf v}\|^2_{{\bf \Sigma}} \equiv {\bf v}^{\top}{\bf \Sigma}^{-1}{\bf v}$. The thermal noise covariance matrix, \mbox{${\bf \Sigma} = {\rm diag}\left(\sigma^2_1,...,\sigma^2_K\right)$}, is determined by each telescope pair antenna parameters~\cite{raymond2021evaluation}. Using Eqs.~\eqref{eq:dynamics}, \eqref{eq:pixel_values}, \eqref{eq:MLP}, each image pixel $p_n$ (elements of $\image$) can be expanded in terms of the unknown parameters:
\begin{equation}
    \pixel_n(\netparams, \rotaxis) = \sum_{\bfx_i \in \raypath_n}  \mlp_{\netparams}\left[ \gamma( \rotation_{\rotaxis} \bfx_i) \right] \intweight_i.
\end{equation}

\section{Experiments}
\label{sec:results}
In this section we describe the simulation setup and show experimental results. These synthetic experiments assume that we are observing SgrA*, the Galactic center black hole that has a mass, $M$, of four million solar masses and zero angular momentum. These intrinsic parameters are necessary to set the magnitude of $\angularvelocity(r)$, as described in Eq.~\eqref{eq:angmom}. Emitting hot-spots are randomly placed roughly at a radius of $1.16 \times r_{\rm ms} $, resulting in an orbital period of $\sim40$ minutes. Each hot-spot is generated as a 3D Gaussian with an isotropic standard deviation of $0.4 \times GM/c^2$.

Ground truth emissions were generated with $128$ temporal frames and $64^3$ 3D voxel grid resolution. The network was implemented in {\tt JAX}~\cite{jax2018github} and optimized using an ADAM optimizer~\cite{kingma2014adam} with a polynomial learning rate transitioning from $1{\rm e}^{-4} \rightarrow 1{\rm e}^{-6}$ over $5{\rm K}$ iterations. EHT measurements were generated using {\tt eht-imaging}~\cite{chael2018interferometric} with realistic thermal noise. Recovery run times were ${\sim} 15$ minutes on two
NVIDIA Titan RTX GPUs. Code implementation is available at the project page~\cite{projectpage}. \vspace{-0.2cm}

\paragraph{Exploring Alternative Emission Representations}
\begin{figure}
	\centering \includegraphics[width=\linewidth]{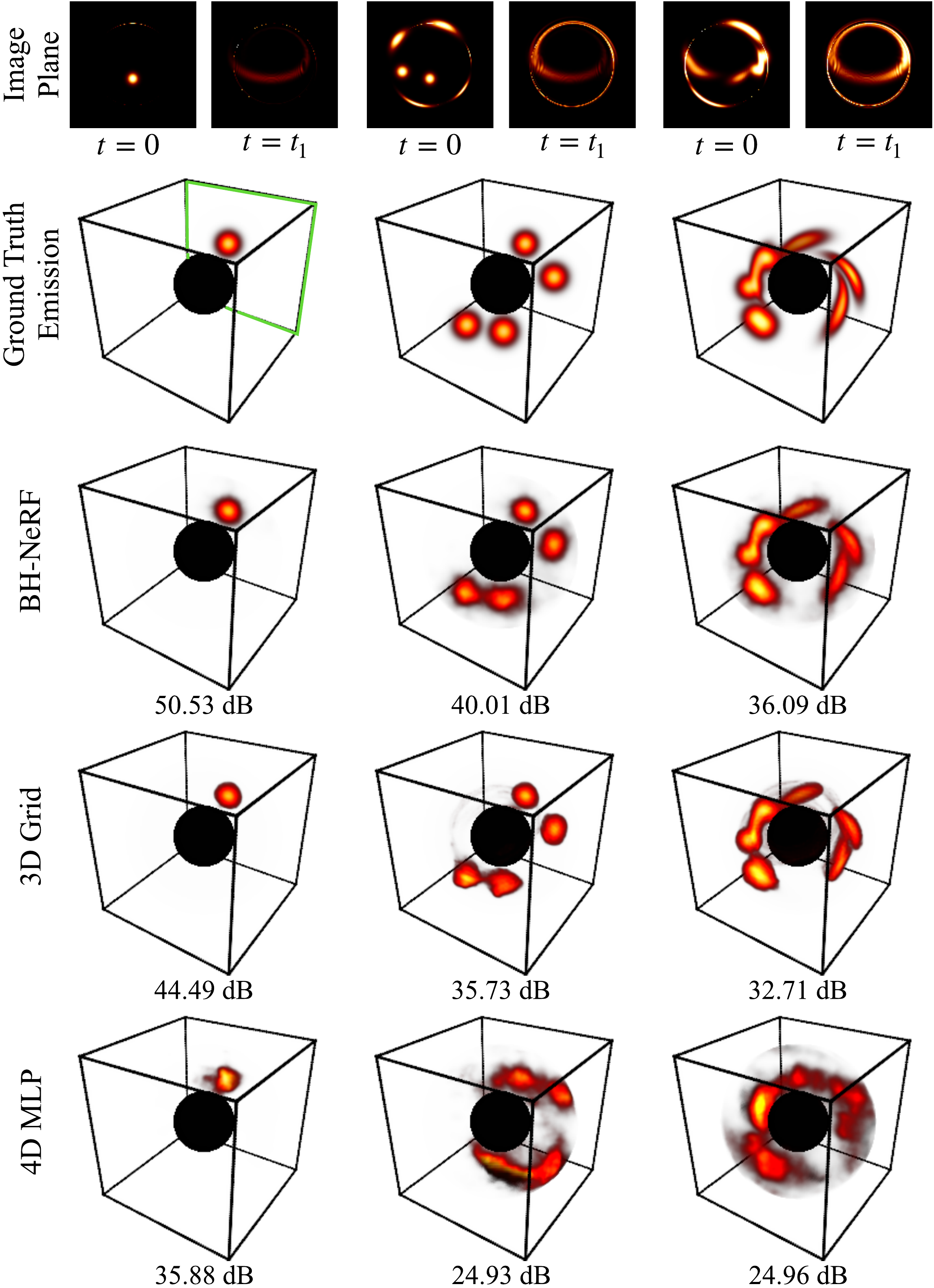}
	\caption{
	Comparison of \approach with two alternative representations. 
	We display the recovered emission for three experimental setups, where the ground truth emission increases in complexity from a single hotspot (left column) to four hotspots (middle column) to eight hotspots (four flaring up before $t=0$).
	In all approaches, we recover emissions from image-domain measurements (projection plane indicated in green) and visualize the emission recovered at time $t=0$. We find that \approach preforms better visually and quantitatively. For quantitative metrics we show the PSNR value comparing recovered emissions the ground truth.
	While \approach significantly outperforms the 4D MLP, we note that this more generalized representation could prove useful in environments with unknown dynamics.
	}
	\label{fig:nerf_vs_grid}
	\vspace{-.2in}
\end{figure}
\begin{figure*}[t]
	\centering \includegraphics[width=\linewidth]{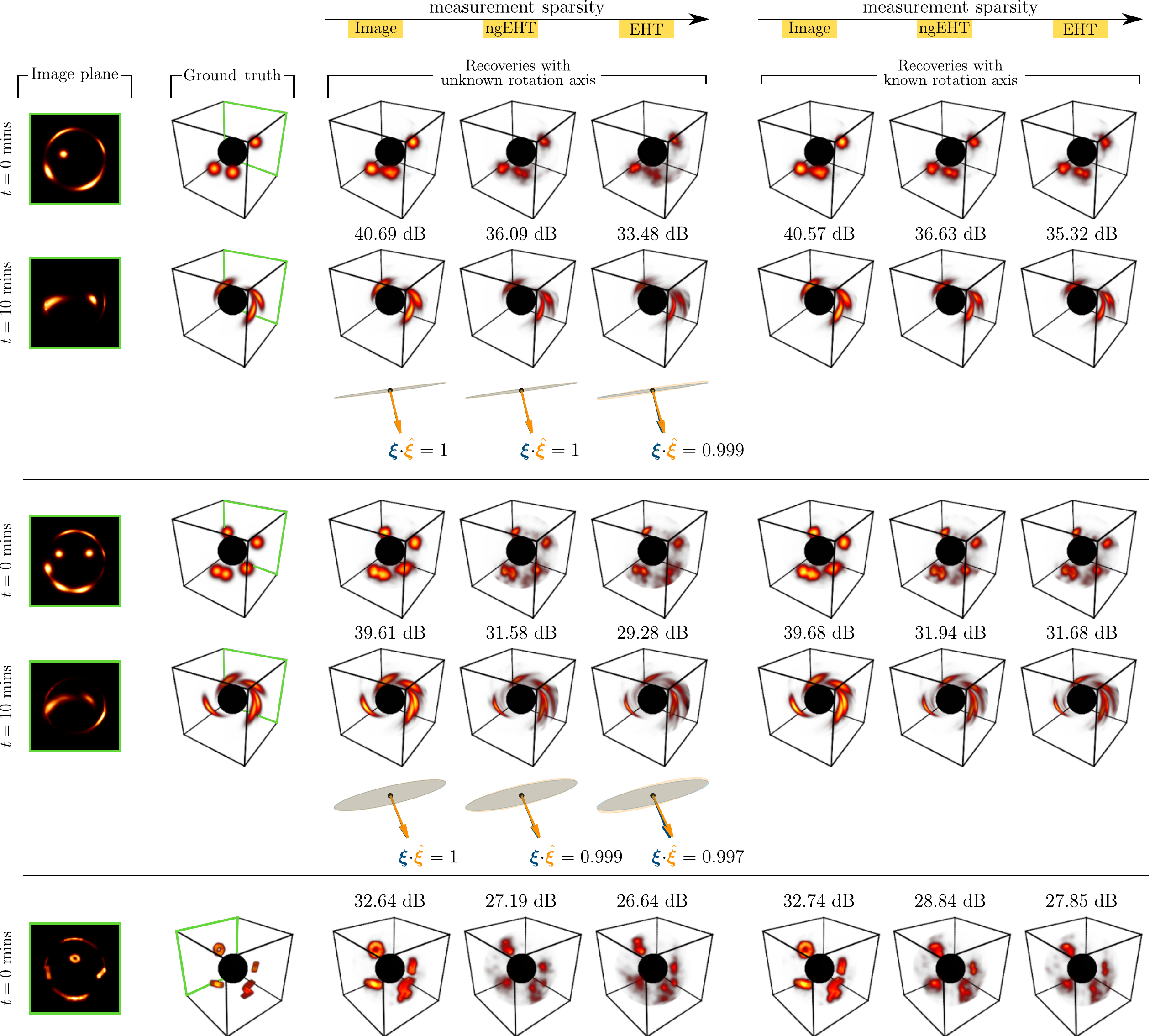}
	\caption{Simulation results for three experiments with different rotation axes and emission patterns. The top two experiments show recoveries of Gaussian emission hotspots at two different times. The third experiment (bottom) shows recovery of 3D digits which illustrates the flexibility of our approach which models arbitrary 3D emission. Each experiment shows: the image-plane projection, ground-truth and recoveries. The orbital period of both is ${\sim}40$ minutes. Green highlights the image-plane where rays are traced from. Recoveries are shown for both unknown (estimated) and known rotation axis. At $t=0$ some of the emission is directly visible and some is ``hidden'' behind the black-hole and is lensed onto the image-plane. Quantitative PSNR values for each recovered emission compared to the ground truth are given below each recovery. For the rotation axis the dot product with the ground truth is indicated below each recovery.}
	\label{fig:sparsity_results}
	\vspace{-.1in}
\end{figure*}

\begin{figure*}[t]
	\centering \includegraphics[width=0.7\linewidth]{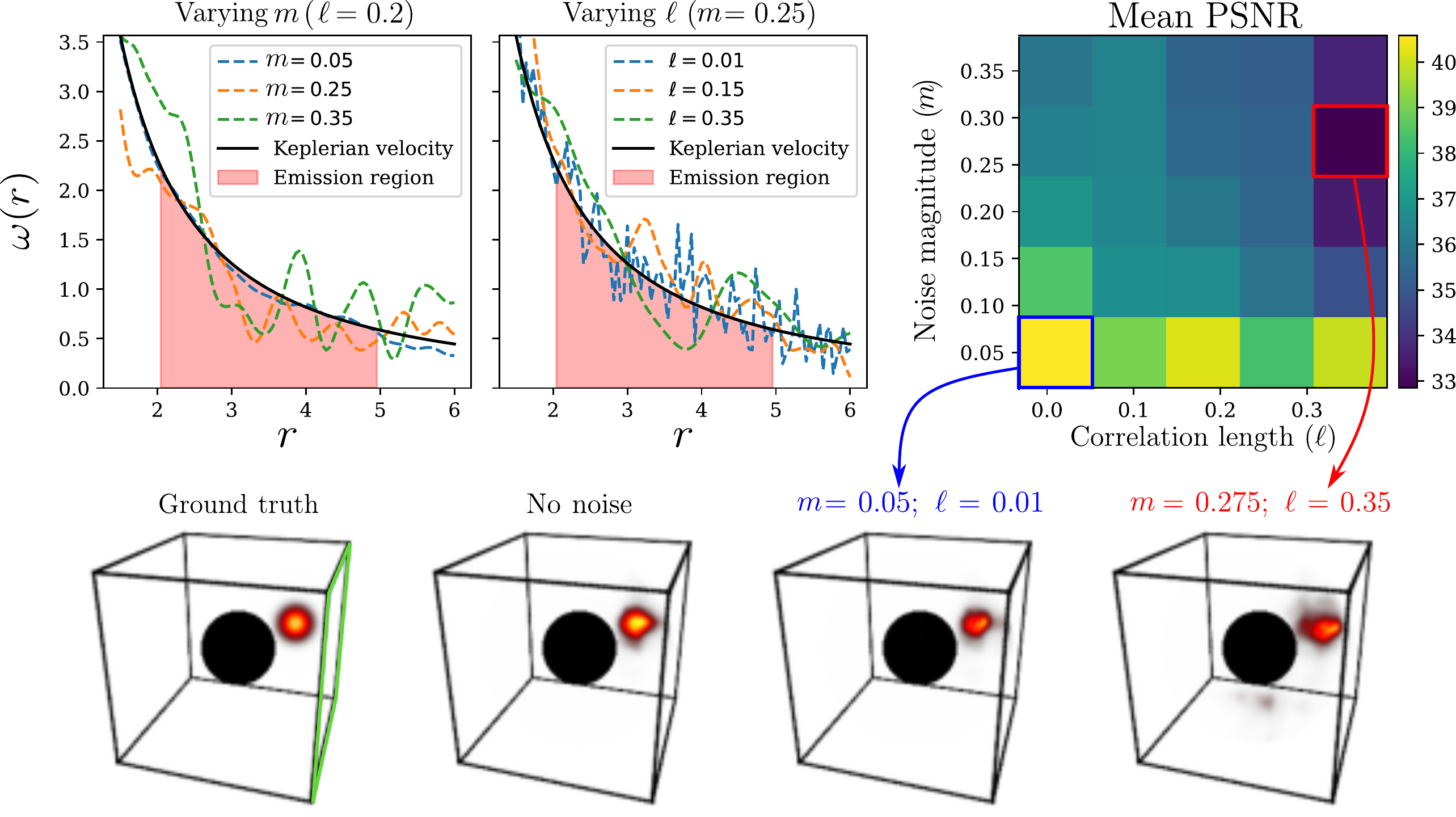}
	\caption{Recovery performance with model mismatch. We perturb the radial velocity model with correlated Gaussian noise. 1) The top panel shows samples of noisy velocity profiles with different noise magnitudes ($m$) and correlation lengths ($\ell$). Red highlights the emission orbit region. 2) The mean PSNR is shown in the top right. 3) The bottom panel shows the volumetric ground-truth, no velocity noise reconstruction, and reconstructions from the highlighted parameter sets. Results are obtained from synthetic ngEHT measurements.}
	\label{fig:v_noise}
	\vspace{-.1in}
\end{figure*}

Our proposed approach, \approach, described in Sec.~\ref{sec:tomography} represents the 4D emission through time as a 3D volume parameterized with a coordinate-based continuous neural network and a parameteric flow field. In the experiments presented in Fig.~\ref{fig:nerf_vs_grid} we explore two alternative representations and compare them to \approach. 

The first approach, which we term {\em 3D Grid}, relies on the Keplerian orbital motion model (described in Sec.~\ref{subsec:emission_dynamics}), however, instead of a continuous neural representation, it uses a discrete voxel grid. The second approach, {\em 4D MLP}, ignores the orbital motion model and recovers a 4D continuous emission field, represented as a MLP similar to \approach (i.e., taking as input a 4D spatiotemproal coordinate rather than a 3D spatial coordinate).


In Fig.~\ref{fig:nerf_vs_grid} we compare emissions recovered with full image-plane measurements and the true rotation axis (not estimated). Results are shown for three different ground truth emission patterns with increasing complexity (from left to right). In the simplest case (left column), a single emission hot-spot flares at $t=0$ in front of the black hole, thus, it is directly visible. The top row of Fig.~\ref{fig:nerf_vs_grid} shows the observations at the initial and final frames: $t=\left\{0, t_1\right\}$. The middle column shows a reconstruction of four hot-spots which flare at $t=0$, where two are directly visible and two are ``hidden'' behind the black-hole and are gravitationally lensed. The last experiment (right column) shows eight hot-spots where four flared up before $t=0$ and four at $t=0$. 

In our experiments, we find that \approach slightly out-performs the 3D voxel grid with minimal tuning, potentially due to its continuous and adaptive allocation of resolution. 
In contrast, the 4D representation preforms significantly worse. This is expected, as it makes no use of any assumption on the emission dynamics. However, we highlight that a 4D representation could be particularly useful in situations where the true dynamics are significantly different from our assumption of Keplarian flow. 

\vspace{-.1in}
\paragraph{Increasing Measurement Sparsity}

In the following simulations, we demonstrate recoveries for three emission patterns orbiting about different axes with known and unknown (estimated) rotation axis. The initial emission in the first experiment (top two rows of Fig.~\ref{fig:sparsity_results}) has three hot-spots with two ``hidden'' behind the black hole and one directly visible in-front of the black hole. The image plane, where rays are traced from, is highlighted in green. For each experiment we show the recovered volume (with PSNR under each reconstruction) and the recovered axis (the dot product of the recovered and true axis specified under the reconstruction). Furthermore, results are shown at $t{=}10~{\rm min}$, with the estimated axis, $\hat{\rotaxis}$, used to advance time. 

A key aspect of \approach is the ability to recover arbitrary emission distributions which enables novel scientific discovery. The third experiment in Fig.~\ref{fig:sparsity_results} demonstrates this ability by recovering 3D MNIST~\cite{mnist3d} digits (`0'--`4') orbiting at a random orientation. Despite the fact that the `2' intially appears behind the black hole, \approach is able to recover its features (further analysis given in the Suppl.~\cite{projectpage}).

While in practice we only have access to sparse visibility measurements of the image plane (Sec.~\ref{subsec:eht}), the image-based recoveries are useful as a form of upper bound (analogous to tiling the Earth with telescopes) on EHT-based results. As expected, increasing the measurements with additional ngEHT telescopes improves the recovery visually and quantitatively over using current EHT telescopes only.

\vspace{-0.1in}
\paragraph{Velocity Model Mismatch}
The Keplerian motion model is a good approximation, however, it does not account for complex dynamics (e.g. due to turbulence or outflowing motion in a magnetically-driven jet \cite{paperV}). To analyze the robustness of our approach to a mismatch in the velocity model we perturb it with correlated Gaussian noise parameterized by magnitude $m$ and correlation length $\ell$. For each parameter set, ($m, \ell$), five velocity profiles were sampled to generate data. Sample reconstructions are shown from the best and worst PSNR parameter sets, highlighted in blue and red respectively. While reconstruction degrades with noise magnitude, our approach is still able to capture the emission structure and estimate the rotation axis with a reasonable level of model mismatch.

\vspace{-0.05in}
\section{Limitations}
While our model is able to recover 3D emission in challenging scenarios with sparse observations, we rely on a few assumptions to do so. First,
we assume knowledge of the black hole's spin and mass in order to pre-compute the curved trajectories of rays and velocity profile. 
Although reasonable ranges of values for these variables can be obtained from other observations, estimating them independently could provide tighter constraints on these scientifically impactful parameters. 
However, estimating these parameters requires back-propagating through the gravitational lensing ray-tracing equations (see supplemental material~\cite{projectpage}). We therefore leave this challenging extension to future work.
Second, we assume a Keplerian dynamics model where the angular velocity is determined by the distance from the black-hole center. Nevertheless, in Sec.~\ref{sec:results} we show that our method is robust to some uncertainty in the velocity profile.
Third, in this work we do not consider additional flares appearing during the observation window. However, as we expect to be able to identify and isolate the time that flares appear by looking for changes in the source's integrated flux (total brightness), we do not expect this to be a significant limitation. 
Fourth, while our experiments include realistic thermal noise, we do not include instrumental errors or particularly challenging atmospheric noise, which would need to be studied before applying our model to real EHT or, in the future, ngEHT measurements.

\section{Conclusion}
This work takes the first steps in showing how ground-based observations from the Event Horizon Telescope could be used to recover evolving 3D emission around the supermassive black hole in our Galactic center. 
Our contributions are two-fold. First, we formulate the novel tomography problem of recovering dynamic 3D emission orbiting a black hole. 
Second, we generalize neural radiance fields to work in scenarios with curved ray trajectories and orbital dynamics induced by the gravitational field around a black hole. This enables the recovery of dynamic 3D emission from sparse measurements of a single viewpoint over time. 

Our simulations and analysis show great promise for tomographic reconstruction of the dynamic environment around a black hole, a direction that can open the door to new insights into our dynamic universe. Moreover, we believe that our work can pave the way for other scientific tomography applications, such as refraction and scattering, which rely on complex ray-tracing at their core.
\section*{Acknowledgements}
AL is supported by the Zuckerman and Viterbi postdoctoral fellowships. AC is supported by Hubble Fellowship grant HST-HF2-51431.001-A awarded by the STScI, operated by the AURA Astronomy for NASA (NAS5-26555). This work was supported by NSF awards 1935980, 2034306, 2048237 and Beyond Limits.

{\small
\bibliographystyle{ieee_fullname}
\bibliography{egbib}
}

\end{document}